\documentclass[a4paper]{article}\usepackage[]{graphicx}\usepackage[]{color}
\makeatletter
\def\maxwidth{ %
  \ifdim\Gin@nat@width>\linewidth
    \linewidth
  \else
    \Gin@nat@width
  \fi
}
\makeatother

\definecolor{fgcolor}{rgb}{0.345, 0.345, 0.345}
\newcommand{\hlnum}[1]{\textcolor[rgb]{0.686,0.059,0.569}{#1}}%
\newcommand{\hlstr}[1]{\textcolor[rgb]{0.192,0.494,0.8}{#1}}%
\newcommand{\hlcom}[1]{\textcolor[rgb]{0.678,0.584,0.686}{\textit{#1}}}%
\newcommand{\hlopt}[1]{\textcolor[rgb]{0,0,0}{#1}}%
\newcommand{\hlstd}[1]{\textcolor[rgb]{0.345,0.345,0.345}{#1}}%
\newcommand{\hlkwb}[1]{\textcolor[rgb]{0.69,0.353,0.396}{#1}}%
\newcommand{\hlkwc}[1]{\textcolor[rgb]{0.333,0.667,0.333}{#1}}%
\newcommand{\hlkwd}[1]{\textcolor[rgb]{0.737,0.353,0.396}{\textbf{#1}}}%

\usepackage{framed}
\makeatletter
\newenvironment{kframe}{%
 \def\at@end@of@kframe{}%
 \ifinner\ifhmode%
  \def\at@end@of@kframe{\end{minipage}}%
  \begin{minipage}{\columnwidth}%
 \fi\fi%
 \def\FrameCommand##1{\hskip\@totalleftmargin \hskip-\fboxsep
 \colorbox{shadecolor}{##1}\hskip-\fboxsep
     \hskip-\linewidth \hskip-\@totalleftmargin \hskip\columnwidth}%
 \MakeFramed {\advance\hsize-\width
   \@totalleftmargin\z@ \linewidth\hsize
   \@setminipage}}%
 {\par\unskip\endMakeFramed%
 \at@end@of@kframe}
\makeatother

\definecolor{shadecolor}{rgb}{.97, .97, .97}
\definecolor{messagecolor}{rgb}{0, 0, 0}
\definecolor{warningcolor}{rgb}{1, 0, 1}
\definecolor{errorcolor}{rgb}{1, 0, 0}
\newenvironment{knitrout}{}{} 

\usepackage{alltt}

\usepackage[a4paper,%
            left=0.5in,right=0.5in,top=1in,bottom=1in]{geometry}

\usepackage[utf8]{inputenc}
\usepackage[T1]{fontenc}
\usepackage{amsthm,amsmath,amssymb,amstext}
\usepackage{booktabs}
\usepackage{array}

\usepackage{nicefrac}
\usepackage{caption}
\usepackage{tabularx}
\usepackage{graphicx}
\usepackage{booktabs}
\usepackage{lscape}
\usepackage{placeins}
\usepackage{colortbl}
\usepackage{dsfont}
\usepackage{multirow}
\usepackage{listings}
\usepackage{adjustbox}
\usepackage{setspace}
\usepackage{url}
\usepackage[round]{natbib}
\usepackage{tikz}
\usetikzlibrary{shapes,arrows}

\IfFileExists{upquote.sty}{\usepackage{upquote}}{}
\IfFileExists{upquote.sty}{\usepackage{upquote}}{}
\begin{document}

\title{Hyperparameters and Tuning Strategies for Random Forest}

\author{by Philipp Probst, Marvin Wright and Anne-Laure Boulesteix}

\maketitle
\abstract{
The random forest algorithm (RF) has several hyperparameters that have to be set by the user, e.g., the number of observations drawn randomly for each tree and whether they are drawn with or without replacement, the number of variables drawn randomly for each split, the splitting rule, the minimum number of samples that a node must contain and the number of trees. 
In this paper, we first provide a literature review on the parameters’ influence on the prediction performance and on variable importance measures.

It is well known that in most cases RF works reasonably well with the default values of the hyperparameters specified in software packages.
Nevertheless, tuning the hyperparameters can improve the performance of RF. In the second part of this paper, after a brief overview of tuning strategies we demonstrate the application of one of the most established tuning strategies, model-based optimization (MBO). 
To make it easier to use, we provide the \textbf{tuneRanger} R package that tunes RF with MBO automatically. 
In a benchmark study on several datasets, we compare the prediction performance and runtime of \textbf{tuneRanger} with other tuning implementations in R and RF with default hyperparameters. 
}

\section{Introduction}

The random forest algorithm (RF) first introduced by \citet{Breiman2001} has now grown to a standard non-parametric classification and regression 
tool for constructing prediction rules based on various types of predictor variables without making any prior assumption on the form of their 
association with the response variable. 
RF has been the topic of several reviews in the last few years including our own review \citep{Boulesteix2012} and others \citep{Criminisi2012, Ziegler2014, Biau2016, Belgiu2016}. 
RF involves several hyperparameters controlling the structure of each individual tree (e.g., the minimal size \textit{nodesize} a node should have to be split) and the structure  and size \color{black} of the forest (e.g., the number of trees) as well as its level of randomness (e.g., the number \textit{mtry} of variables considered as candidate splitting variables at each split or the sampling scheme used to generate the datasets on which the trees are built). 
The impact of these hyperparameters has been studied in a number of papers. 
However, results on this impact are often focused on single hyperparameters and provided as by-product of studies devoted to other topics (e.g., a new variant of RF) and thus are difficult to find for readers without profound knowledge of the literature. Clear guidance is missing and the choice of adequate values for the parameters remains a challenge in practice. 

It is important to note that RF  may be used \color{black} in practice for two different purposes. In some RF applications, the focus is on the construction of a classification or regression rule with good accuracy that is intended to be used as a prediction tool on future data. In this case, the objective is to derive a rule with high prediction performance---where performance can be defined in different ways depending on the context, the simplest approach being to consider the classification error rate in the case of classification and the mean squared error in the case of regression. In other RF applications, however, the goal is not to derive a classification or regression rule but 
to investigate the relevance of the candidate predictor variables for the prediction problem at hand or, in other words, to assess their respective contribution to the prediction of the response variable. See the discussion by \cite{shmueli2010explain} on the difference between \lq\lq predicting'' and \lq\lq explaining''. These two objectives have to be kept in mind when investigating the effect of parameters.  Note, however, that there might be overlap of these two objectives: For example one might use a variable selection procedure based on variable importance measures to obtain a well performing prediction rules using RF.

Note that most hyperparameters are so-called \lq\lq tuning parameters'', in the sense that their values have to be optimized carefully---because the optimal values are dependent on the dataset at hand. 
 Optimality here refers to a certain performance measure that have to be chosen beforehand. An important concept related to parameter tuning is overfitting: parameter values corresponding to complex rules tend to {\it overfit} the training data, i.e. to yield prediction rules that are too specific to the training data---and perform very well for this data but probably worse for independent data. The selection of such sub-optimal parameter values can be partly avoided by using a test dataset or cross-validation procedures for tuning. In the case of random forest the out-of-bag observations can also be used. \color{black}

We will see that for random forest not all but most presented parameters are tuning parameters. 
Furthermore, note that the distinction between hyperparameters and algorithm variants is blurred. 
For example, the splitting rule may be considered as a (categorical) hyperparameter, but also as defining distinct variants of the RF algorithm. 
Some arbitrariness is unavoidable when distinguishing hyperparameters from variants of RF. 
In the present paper, considered hyperparameters are the number of candidate variables considered at each split (commonly denoted as \textit{mtry}), the hyperparameters specifying the sampling scheme (the \textit{replace} argument and the sample size), the minimal node size and related parameters, the number of trees, and the splitting rule.

This paper addresses specifically the problem of the choice of parameters of the random forest algorithm from two different perspectives. Its first part presents a review of the literature on the choice of the various parameters of RF, while the second part presents different tuning strategies and software
packages for obtaining optimal hyperparameter values which are finally compared in a benchmark study. 

  \section{Literature Review}
  \label{sec:literature}
  
In the first section of this literature review, we focus on the influence of the hyperparameters on the prediction performance, e.g., the error rate or the  area under the ROC Curve (AUC), \color{black} and the runtime of random forest, while literature dealing specifically with the influence on the variable importance is reviewed in the second section. In Table \ref{hyperpar_table} the different hyperparameters with description and typical default values are displayed. 

\begin{table}[ht]
\centering
\begin{tabular}{llr}
  \hline
 Hyperparameter & Description & Typical default values\\ 
  \hline
  mtry & Number of drawn candidate variables in each split & $\sqrt{p}$, $p/3$ for regression \\ 
  sample size & Number of observations that are drawn for each tree & $n$ \\
  replacement & Draw observations with or without replacement & TRUE (with replacement)\\
  node size & Minimum number of observations in a terminal node & 1 for classification, 5 for regression \\
  number of trees & Number of trees in the forest & 500, 1000 \\
  splitting rule & Splitting criteria in the nodes & Gini impurity, $p$-value, random\\
   \hline
\end{tabular}
\caption{Overview of the different hyperparameter of random forest and typical default values. $n$ is the number of observations and $p$ is the number of variables in the dataset. 
}
\label{hyperpar_table}
\end{table}
\color{black}

  \subsection{Influence on performance}
  \label{sec:lit_param}
  
As outlined in \citet{Breiman2001}, \textit{\lq\lq [t]he randomness used in tree construction has to aim for low correlation $\rho$ while maintaining reasonable strength''}. 
In other words, an optimal compromise between low correlation and reasonable strength of the trees has to be found. 
This can be controlled by the parameters \textit{mtry}, sample size and node size which will be presented in Section~\ref{sec:mtry}, \ref{sec:sampsize} and \ref{sec:nodesize}, respectively. Section~\ref{sec:ntree} handles the number of trees, while Section~\ref{sec:splitrule} is devoted to the splitting criterion.

  \subsubsection{Number of randomly drawn candidate variables (\textit{mtry})}
  \label{sec:mtry}
One of the central hyperparameters of RF is \textit{mtry}, as denoted in most RF packages, which is defined as the number of randomly drawn candidate variables out of which each split is selected when growing a tree. Lower values  of \textit{mtry}  lead to more different, less correlated trees, yielding better stability when aggregating. Forests constructed with a low \textit{mtry} also tend to better exploit variables with moderate effect on the response variable, that would be masked by variables with strong effect if those had been candidates for splitting. However, lower values of \textit{mtry} also lead to trees that perform on average worse, since they are built based on suboptimal variables (that were selected out of a small set of randomly drawn candidates): possibly non-important variables are chosen. We have to deal with a trade-off between stability and accuracy of the single trees.

As default value in several software packages \textit{mtry} is set to $\sqrt{p}$ for classification and  $p/3$ for regression with $p$ being the number of predictor variables. In their paper on the influence of hyperparameters on the accuracy of RF, \citet{Bernard2009} conclude that $\textit{mtry} = \sqrt{p}$ is a reasonable value, but can sometimes be improved. They especially outline that the real number of relevant predictor variables highly influences the optimal \textit{mtry}. If there are many relevant predictor variables, \textit{mtry} should be set small because then not only the strongest influential variables are chosen in the splits but also less influential variables, which can provide small but relevant performance gains. These less influential variables might, for example, be useful for the prediction of a small group of observations that stronger variables fail to predict correctly.  If \textit{mtry} is large, however, these less influential variables might not have the chance to contribute to prediction because  stronger variables are preferably selected for splitting and thus \lq\lq mask'' the smaller effects. On the other hand, \color{black} if there are only a few relevant variables out of many, which is the case in many genetic datasets, \textit{mtry} should be set high, so that the algorithm can find the relevant variables \citep{Goldstein2011}.  A large {\it mtry} ensures that there is (with high probability) at least one strong variable in the set of {\it mtry} candidate variables.\color{black} 
   
Further empirical results are provided by \cite{Genuer2008}. In their low dimensional classification problems $\textit{mtry} = \sqrt{p}$ is
convenient regarding the error rate. For low dimensional regression problems, 
in their examples $\sqrt{p}$ performs better than $p/3$ regarding the mean squared error. For high dimensional data they observe lower error 
rates for higher \textit{mtry} values for both classification and regression, corroborating \cite{Goldstein2011}.
  
Computation time decreases approximately linearly with lower \textit{mtry} values \citep{Wright2017}, since most of RF's computing time is devoted to the selection of the split variables. 

  \subsubsection{Sampling scheme: sample size and replacement}
  \label{sec:sampsize}
  
  The sample size parameter determines how many observations are drawn for the training of each tree. 
  It has a similar effect as the \textit{mtry} parameter. Decreasing the sample size leads to more diverse trees and thereby lower correlation between the trees, which has a positive effect on the prediction accuracy \color{black}when aggregating the trees. However, the accuracy \color{black}of the single trees decreases, since fewer observations are used for training. Hence, similarly to the \textit{mtry} parameter, the choice of the sample size can be seen as a trade-off between stability and accuracy of the trees. \citet{Martinez2010} carried out an empirical analysis of the dependence of the performance on the sample size. They concluded that the optimal value is
problem dependent and can be estimated with the out-of-bag predictions. In most datasets they observed better performances when sampling less observations 
than the standard choice (which is to sample as many observations with replacement as the number of observations in the dataset). Setting it to lower values
reduces the runtime.

Moreover, \citet{Martinez2010} claim that there is no substantial performance difference between sampling with replacement or without replacement when the sample size parameter is set optimally. However, both theoretical \citep{janitza2016pitfalls} and empirical results \citep{Strobl2007} show that sampling with replacement may induce a slight variable selection bias when categorical variables with varying number of categories are considered. In these specific cases, performance may be impaired by sampling {\it with} replacement, even if this impairment could not be observed by \cite{Martinez2010} when considering averages over datasets of different types. 

\subsubsection{Node size}
  \label{sec:nodesize}
  The \textit{nodesize} parameter specifies the minimum number of observations in a terminal node. Setting it lower leads to trees with a larger depth which means that more splits are performed until the terminal nodes. In several standard software packages the default value is 1 for classification and 5  for regression. It is believed to generally provide good results \citep{Uriarte2006, Goldstein2011} but performance can potentially be improved by tuning it \citep{Lin2006}. In particular, \cite{Segal2004} showed an example where increasing the number of noise variables leads to a higher optimal node size. 
  
Our own preliminary experiments suggest that the computation time decreases approximately exponentially with increasing node size. In our experience, especially in large sample datasets it may be helpful to set this parameter to a value higher than the default as it decreases the runtime substantially, often without substantial loss of prediction performance \citep{Segal2004}.
  
Note that other hyperparameters than the node size may be considered to control the size of the trees. For example, the R package 
\textbf{party} \citep{hothorn2006unbiased} allows to set the minimal size, \textit{minbucket}, that child nodes should have for the split to be performed. The hyperparameters \textit{nodesize} and \textit{minbucket} are obviously related, since the size of all parent nodes equals at least twice the value of \textit{minbucket}. However, setting \textit{minbucket} to a certain value does not in general lead to the same trees as setting \textit{nodesize} to double this value. To explain this, let us consider a node of size $n=10$ and a candidate categorical predictor variable taking value 1 for $n_1=9$ of the $n=10$ observations of the node, and value 0 for the remaining observation. If we set \textit{nodesize} to 10 and do not put any restriction on \textit{minbucket} (i.e., set it to 1), our candidate variable can be selected for splitting. If, however, we proceed the other way around and set \textit{minbucket} to 5 while not putting any restriction on \textit{nodesize} (i.e., while setting it to 2), our candidate variable cannot be selected, because it would produce a---too small---child node of size 1. On the one hand, one may argue that splits with two large enough child nodes are preferable---an argument in favor of setting \textit{minbucket} to a value larger than one. On the other hand, this may yield a selection bias in the case of categorical variables, as demonstrated through this simple example and also discussed in \cite{boulesteix2012random} in the context of genetic data.

Furthermore, in the R package \textbf{randomForest} \citep{Liaw2002}, it is possible to specify \textit{maxnodes}, the maximum number of terminal nodes 
that trees in the forest can have, while the R package \textbf{party} allows to specify the strongly related hyperparameter 
\textit{maxdepth}, the maximal depth of the trees, which is the maximum number of splits until the terminal node.


  \subsubsection{Number of trees}
  \label{sec:ntree}
  
  The number of trees in a forest is a parameter that is not tunable in the classical sense but should be set sufficiently 
  high \citep{Uriarte2006, Oshiro2012, Probst2017}. Out-of-bag error curves (slightly) increasing with the number of trees are occasionally observed for certain error measures \citep[see][for an empirical study based on a large number of datasets]{Probst2017}. According to measures based on the mean quadratic loss such as the mean squared error (in case of regression) or the Brier score (in case of classification), however, more trees are always better, as theoretically proved by \citet{Probst2017}.
  
 The convergence rate, and thus the number of trees needed to obtain optimal performance, depends on the dataset's properties. Using a large number of datasets, \citet{Oshiro2012} and \citet{Probst2017} show empirically that the biggest performance gain can often be achieved when growing the first 100 trees. 
The convergence behaviour can be investigated by inspecting the out-of-bag curves showing the performance for a growing number of trees. 
\citet{Probst2017} argue that the error rate is not the optimal measure for that purpose  because, by considering a prediction as either true or false, one ignores much of the information output by the RF and focuses too much on observations that are close to the prediction boundary. \color{black} Instead, they recommend the use of other measures based on the predicted class probabilities such as the Brier score or the logarithmic loss, as implemented in the R package \textbf{OOBCurve} \citep{Probst20172}.
  
Note that the convergence rate of RF does not only depend on the considered dataset's characteristics but possibly also on  hyperparameters. Lower sample size (see Section~\ref{sec:sampsize}), higher node size values (see Section~\ref{sec:nodesize}) and smaller \textit{mtry} values (see Section~\ref{sec:mtry}) lead to less correlated trees.  These trees are more different from each other and are expected to provide more different predictions. Therefore, we suppose that more trees are needed to get clear predictions for each observation which leads to a higher number of trees for obtaining convergence. \color{black}

  
  \newpage
 The computation time increases linearly with the number of trees. As trees are trained independently from each other they can be trained in parallel on several CPU cores which is implemented in software packages such as \textbf{ranger} \citep{Wright2017}.
  
\subsubsection{Splitting rule}
  \label{sec:splitrule}
  
  The splitting rule is not a classical hyperparameter as it can be seen as one of the core properties characterizing the RF. However, it can in a large sense also be considered as a categorical hyperparameter.  
  The default splitting rule of Breiman's original RF \citep{Breiman2001} consists of selecting, out of all splits of the (randomly selected \textit{mtry}) candidate variables, the split that minimizes the Gini impurity (in the case of classification) and the weighted variance (in case of regression). This method favors the selection of variables with many possible splits (e.g., continuous variables or categorical variables with many categories) over variables with few splits (the extreme case being binary variables, which have only one possible split) due to multiple testing mechanisms \citep{Strobl2007}. 
  
Conditional inference forests (CIF) introduced by \cite{hothorn2006unbiased} and implemented in the R package \textbf{party} and in the newer package \textbf{partykit} \citep{Hothorn2015} allow to avoid this variable selection bias by selecting the variable with the smallest $p$-value in a global test (i.e., {\it without} assessing all possible splits successively) in a first step, and selecting the best split from the selected variable in a second step  by maximizing a linear test statistic. \color{black} Note that the global test to be used in the first step depends on the scale of the predictor variables and response variable. \cite{hothorn2006unbiased} suggest several variants. A computationally fast alternative using $p$-value approximations for maximally selected rank statistics is proposed by \cite{Wright2017b}. This variant is available in the \textbf{ranger} package for regression and survival outcomes.

When tests are performed for split selection, it may only make sense to split if the $p$-values fall below a certain threshold, which should then be considered as a hyperparameter. In the R package \textbf{party} the hyperparameter \textit{mincriterion} represents one minus the $p$-value threshold and in \textbf{ranger} the hyperparameter \textit{alpha} is the $p$-value threshold.

To increase computational efficiency, splitting rules can be randomized \citep{Geurts2006}. To this end, only a randomly selected subset of possible splitting values are considered for a variable. The size of these subsets is specified by the hyperparameter \textit{numRandomCuts} in the \textbf{extraTrees} package and by \textit{num.random.splits} in \textbf{ranger}. If this value is set to $1$, this variant is called \textit{extremely randomized trees} \citep{Geurts2006}. In addition to runtime reduction, randomized splitting might also be used to add a further component of randomness to the trees, similar to \textit{mtry} and the sample size.

Until now, none of the existing splitting rules could be proven as superior to the others in general regarding the performance. For example, the splitting rule based on the decrease of Gini impurity implemented in Breiman's original version is affected by a serious variable selection bias as outlined above. However, if one considers datasets in which variables with many categories are more informative than variables with less categories, this variable selection bias---even if it is in principle a flaw of the approach---may accidentally lead to improved accuracy. Hence, depending on the dataset and its properties one or the other method may be better. Appropriate benchmark studies based on simulation or real data have to be designed, to evaluate in which situations which splitting rule performs better. 
\color{black}

Regarding runtime, extremely randomized trees are the fastest as the cutpoints are drawn completely randomly, followed by the classical random forest, while for conditional inference forests the runtime is the largest. 

  \subsection{Influence on variable importance}
  
  The RF variable importance \citep{Breiman2001} is a measure reflecting the importance of a variable in a RF prediction rule. \color{black} 
  While effect sizes and p-values of the Wald test or likelihood ratio tests are often used to assess the importance of variables in case of logistic or 
  linear regression, the RF variable importance measure can also  automatically \color{black} capture non-linear and interaction effects  without specifying these a priori \citep{Wright2016} \color{black} and is also applicable when more variables than observations are available. Several variants of the variable importance exist, including the Gini variable importance measure and the permutation variable importance measure \citep{Breiman2001,Strobl2007}. In this section we focus on the latter,  since the former has been shown to be strongly biased.  The Gini variable importance measure assigns higher importance values to variables with more categories or continuous variables \citep{Strobl2007} and to categorical variables with equally sized categories \citep{boulesteix2012random} even if all variables are independent of the response variable. \color{black}
  
Many of the effects of the hyperparameters described in the previous section  \ref{sec:lit_param} are expected to also have an effect on the variable importance. However, specific research on this influence is still in its infancy. 
 Most extensive is probably the research about the influence of the number of trees. In contrast, the literature is very scarce as far as the sample size 
and node size are concerned. 
  
  \subsubsection{Number of trees}
  
   More trees are generally required for stable variable importance estimates \citep{Genuer2010, Goldstein2011}, than for the simple prediction 
purpose. \citet{Lunetta2004} performed simulations with more noisy variables than truly associated covariates 
  and concluded that multiple thousands of trees must be trained in order to get stable estimates of the 
  variable importance. The more trees are trained, the more stable the predictions should be for the 
  variable importance. In order to assess the stability one could train several random forests with a 
  fixed number of trees and check whether the ranking of the variables by importance are different between
  the forests. 
  
  \subsubsection{\textit{mtry}, splitting rule and node size}
 \citet{Genuer2010} examine the influence of the parameter \textit{mtry} on the variable importance. They conclude that increasing the \textit{mtry} value leads to much higher magnitudes of the variable importances. As already outlined in Section~\ref{sec:splitrule} the random forest standard splitting rule is biased when predictor variables vary in 
their scale. This has also a substantial impact on the variable importance \citep{Strobl2007}. In the case of the Gini variable importance, 
predictor variables with many categories or numerical values receive on average a higher variable importance than binary variables if both variables have no influence on the outcome variable. The permutation variable 
importance remains unbiased in these cases, but there is a higher variance of the variable importance for variables with many categories. 
This could not be observed for conditional inference forests combined with subsampling (without replacement) as sampling procedure and
therefore \citet{Strobl2007} recommend to use this method for getting reliable variable importance measures.

  \citet{Groemping2009} compared the influence of \textit{mtry} and the node size on the variable importance of the 
  standard random forest and of the conditional inference forest. 
  Higher \textit{mtry} values lead to lower variable importance of weak regressors.
  The values of the variable importance from the standard random forest were far less dependent on \textit{mtry}
  than the ones from the conditional inference forests. This was due to the much larger size (i.e., number of splits until the terminal node)
  of individual trees in the standard random forest. Decreasing the tree size (for example 
  by setting a higher node size value) while setting \textit{mtry} to a small value leads to more equal values of  
  the variable importances of all variables, because there was less chance that relevant variables were chosen in the splitting procedures. 
  

\section{Tuning random forest}

Tuning is the task of finding optimal hyperparameters for a learning algorithm for a considered dataset. In supervised learning (e.g., regression and classification), optimality may refer to different performance measures (e.g., the error rate or the AUC)  and to \color{black} the runtime which can highly depend on hyperparameters in some cases as outlined in Section~ \ref{sec:literature}. In this paper we mainly focus on the optimality regarding performance measures. 

Even if the choice of adequate values of hyperparameters has been partially investigated in a number of studies as reviewed in Section~2,   unsurprisingly \color{black} the literature provides general trends rather than clear-cut guidance. In practice, users of RF are often unsure whether alternative values of tuning parameters may improve performance compared to default values.  Default values are given by software packages or can be calculated by using previous datasets \citep{Probst2018}\color{black}. In the following section we will review literature about  
the \lq\lq tunability'' of random forest. \lq\lq Tunability'' is defined as the amount of performance gain  compared to default values that can 
be achieved by tuning  one hyperparameter  (\lq\lq tunability'' of the hyperparameter) or all hyperparameters (\lq\lq tunability'' of the algorithm); see \cite{Probst2018} for more details. \color{black} Afterwards, evaluation strategies, evaluation measures and tuning search strategies are presented.
Then we review software implementations of tuning of RF in the programming language R and finally show the results of a large scale benchmark study comparing the different implementations.

\subsection{Tunability of random forest}

  Random forest is an algorithm which is known to provide good results in the default settings \citep{Delgado2014}. \citet{Probst2018} measure the \lq\lq tunability'' of
algorithms and hyperparameters of algorithms and conclude that random forest is far less tunable than other algorithms such as  support vector machines. Nevertheless, a small performance gain (e.g., an average increase of the AUC of  0.010 \color{black} based on the 38 considered datasets) can be achieved via tuning compared to the  default software package \color{black} hyperparameter values. This average performance gain, although moderate, can be an important improvement in some cases,  when, for example, each wrongly classified observation implies high costs. Moreover, for some datasets it is much higher than 0.01 (e.g., around 0.03). \color{black}

  
  As outlined in Section~\ref{sec:literature}, all considered hyperparameters might have an effect on the performance of RF. It is not completely clear, however, which of them should routinely be tuned in practice. Beyond the special case of RF, \citet{Probst2018} suggest a general framework to assess the tunability of different hyperparameters of different algorithms (including RF) and illustrate their approach through an application to 38 datasets. In their study, tuning the parameter \textit{mtry} provides the biggest average improvement of the AUC (0.006), followed by the sample size (0.004), while the node size had only a small effect (0.001). Changing the 
  \textit{replace} parameter from drawing with replacement to drawing without 
replacement also had a small positive effect (0.002). Similar results were observed in the work of \citet{van2018hyperparameter}. \color{black}As outlined in Section \ref{sec:ntree}, the number of trees cannot be seen as tuning parameter: higher values are generally preferable to smaller values with respect to performance.
If the performance of RF with default values of the hyperparameters can be improved by choosing other values, the next question is how this choice should be performed. 

\subsection{Evaluation strategies and evaluation measures}

A typical strategy to evaluate the performance of an algorithm with different values of the hyperparameters in the context of tuning is $k$-fold cross-validation. The number $k$ of folds is usually chosen between 2 and 10. Averaging the results of several repetitions of the whole cross-validation procedure provides more reliable results as the variance of the estimation is reduced \citep{seibold2018choice}.
  
In RF (or in general when bagging is used) another strategy is possible, namely using the out-of-bag observations to evaluate the trained algorithm. Generally, the results of this strategy are reliable \citep{Breiman1996}, i.e., approximate the performance of the RF on independent data reasonably well. A bias can be observed in special data situations \citep[see][and references therein]{janitza2018overestimation}, for example in very small datasets with $n<20$, when there are many predictor variables and balanced classes. Since these problems are specific to particular and relatively rare situations and tuning based on out-of-bag-predictions has a much smaller runtime than procedures such as $k$-fold cross-validation (which is especially important in big datasets), we recommend the out-of-bag approach for tuning  as an appropriate procedure for most datasets\color{black}.  
 
The evaluation measure is a measure that is dependent on the learning problem. In classification, two of the most commonly considered measures are the classification error rate and the Area Under the Curve (AUC). Two other common measures that are based on probabilities are the Brier score and the logarithmic loss. An overview of evaluation measures for classification is given in \citet{Ferri2009}. 

\subsection{Tuning search strategies}
  
Search strategies differ in the way the candidate hyperparameter values (i.e., the values that have to be evaluated with respect to their out-of-bag performance) are chosen. Some strategies specify all the candidate hyperparameter values from the beginning, for example, random search and grid search presented in the following subsections. In contrast, other more sophisticated methods such as F-Race \citep{Birattari2010}, general simulated annealing \citep{Bohachevsky1986} or  sequential model-based optimization (SMBO) \citep{Jones1998, Hutter2011} iteratively use the results of the different already evaluated hyperparameter values and choose future hyperparameters considering these results. 
The latter procedure, SMBO, is introduced at the end of this section and used in two of the software implementations that are presented in the sections 
\ref{sec:existtune} and \ref{sec:tuneRanger}. 
  
  \subsubsection*{Grid search and random search}
  \label{sec:gridsearch}
 One of the simplest strategies is grid search, in which all possible combinations of given discrete parameter spaces are evaluated. Continuous parameters have to be discretized beforehand. Another approach is random search, in which hyperparameter values are drawn randomly 
  (e.g., from a uniform distribution) from a specified hyperparameter space. \citet{Bergstra2012} show that for neural networks random search is more efficient in searching good hyperparameter specifications than grid search.

  \subsubsection*{Sequential model-based optimization}
   \label{sec:mbo}
  Sequential model-based optimization (SMBO) is a very successful tuning strategy that iteratively tries to find the best hyperparameter settings
based on evaluations of hyperparameters that were done beforehand. SMBO is grounded in the \lq\lq black-box function optimization'' literature \citep{Jones1998} and achieves state-of-the-art performance for solving a number of optimization problems \citep{Hutter2011}. We shortly describe the SMBO algorithm implemented in the R package \textbf{mlrMBO} \citep{Bischl20171}, which is also used in the R package \textbf{tuneRanger} \citep{Probst20182} described in Section \ref{sec:tuneRanger}. It consists of the following steps:
  
  \begin{enumerate}
  \item Specify an evaluation measure (e.g., the AUC in the case of classification or the mean squared error in the case of regression), also sometimes denoted as \lq\lq target outcome'' in the literature,  an evaluation strategy (e.g., 5-fold cross-validation) \color{black} and a constrained hyperparameter space on which the tuning should be executed. 
  \item Create an initial design, i.e., draw random points from the hyperparameter space and evaluate them 
  (i.e., evaluate the chosen evaluation measure using the chosen evaluation strategy).
  \item Based on the results obtained from the previous step following steps are iteratively repeated:
  \begin{enumerate}
  \item Fit a regression model (also called surrogate model, e.g., kriging \citep{Jones1998} or RF) based on all already evaluated design points with the evaluation measure as the dependent variable and the hyperparameters as predictor variables.
  \item Propose a new point for evaluation on the hyperparameter space based on an infill criterion. This criterion is based on the surrogate 
  model and proposes points that have good expected outcome values and lie in regions of the hyperparameter space where not many points were evaluated yet.
  \item Evaluate the point and add it to the already existing design points. 
  \end{enumerate}
  \end{enumerate}
  
    \subsection{Existing software implementations}
   \label{sec:existtune}
Several packages already implement such automatic tuning procedures for RF. We shortly describe the three most common ones in R:
  \begin{itemize}
 \item \textbf{mlrHyperopt} \citep{Richter2017} uses sequential model-based optimization as implemented in the R package \textbf{mlrMBO}. It has predefined tuning parameters and tuning spaces for many supervised learning algorithms that are part of the \textbf{mlr} package. 
Only one line of code is needed to perform the tuning of these algorithms with the package. 
In case of \textbf{ranger}, the default parameters that are tuned are \textit{mtry} (between 1 and the number of variables $p$) and the node size (from 1 to 10), with 25 iteration steps (step number 3 in the previous subsection) and no initial design. In \textbf{mlrHyperopt}, the standard evaluation strategy in each iteration step of the tuning procedure 
is 10-fold cross-validation and the evaluation measure is the mean missclassification error. The parameters, the evaluation strategy and measure can be 
  changed by the user. As it is intended as a platform for sharing tuning parameters and spaces, users can use their own tuning parameters and spaces and upload them to the webservice of the package. 
  \item \textbf{caret} \citep{Kuhn2008} is a set of functions that attempts to streamline the process for creating predictive models. When 
 executing \textbf{ranger} via \textbf{caret} it automatically performs a grid search of \textit{mtry} over 
 the whole \textit{mtry} parameter space.
 By default, the algorithm evaluates 3 points in the parameter space (smallest and biggest possible mtry and the mean of these two values) with 25 bootstrap iterations as evaluation strategy. 
 The algorithm finally chooses the value with the lowest error rate in case of classification and the lowest mean squared error in case of regression. 
 \item \textbf{\texttt{tuneRF}} from the \textbf{randomForest} package implements an automatic tuning procedure for the \textit{mtry} parameter. 
First, it calculates the out-of-bag error with the default \textit{mtry} value (square root of the number of variables $p$ for classification and $p/3$ for regression). Second, it tries out a new smaller value of \textit{mtry} (default is to deflate \textit{mtry} by the factor 2). If it provides a better  out-of-bag error rate (relative improvement of at least 0.05) \color{black} the algorithm continues trying out smaller \textit{mtry} values in the same way. After that, the algorithm tries out larger values than the default of \textit{mtry} until there is no more improvement, analogously to the second step. Finally the algorithm returns the model with the best \textit{mtry} value.
\end{itemize}

\subsection{The \textbf{tuneRanger} package}
 \label{sec:tuneRanger}
As a by-product of our literature review on tuning for RF, we created a package, \textbf{tuneRanger} \citep{Probst20182}, for automatic tuning of RF based on the package \textbf{ranger} through a single line of code, implementing all features that we identified as useful in other packages, and intended for users who are not very familiar with tuning strategies. The package \textbf{tuneRanger}  is mainly based on the R packages \textbf{ranger} \citep{Wright2017}, \textbf{mlrMBO} \citep{Bischl20171} and \textbf{mlr} \citep{Bischl2016}. 
  The main function \texttt{tuneRanger} of the package works internally as follows:
  \begin{itemize}
  \item Sequential model-based optimization (SMBO, see Section~\ref{sec:mbo}) is used as tuning strategy with 30 evaluated random points for the initial design and 
70 iterative steps in the optimization procedure. The number of  steps for the initial design and in the optimization procedure  are \color{black} parameters that can be changed by the user, although the default settings  30 resp. 70 \color{black} provide good results in our experiments. 
  \item As a default, the function simultaneously tunes the three parameters \textit{mtry}, sample size and node size. 
  \textit{mtry} values are sampled from the space $[0,p]$ with $p$ being the number of predictor variables, while sample size values are sampled from $[0.2 \cdot n, 0.9 \cdot n]$ with $n$ being the number of observations. Node size values are sampled with higher probability (in the initial design) for smaller values by sampling x from $[0,1]$ and transforming the value by the formula $[(n \cdot 0.2)^x]$. 
The tuned parameters can be changed by the user by changing the argument \texttt{tune.parameters}, if, for example, only the \textit{mtry} value should be tuned  or if additional parameters, such as the sampling strategy (sampling with or without resampling) or the handling of unordered factors (see \cite{Hastie2001}, chapter 9.2.4 or the help of the \texttt{ranger} package for more details), should be included in the tuning process. \color{black}
  \item Out-of-bag predictions are used for evaluation, which makes it much faster than other packages that use evaluation strategies such as 
  cross-validation. 
  \item Classification as well as regression is supported.
 \item The default measure that is optimized is the Brier score for classification, which yields a finer evaluation than the commonly used error rate \citep{Probst2017}, and the mean squared error for regression. 
  It can be changed to any of the 50 measures currently implemented in the R package \textbf{mlr} and documented in the online tutorial \citep{Schiffner2016}: \url{http://mlr-org.github.io/mlr-tutorial/devel/html/measures/index.html}
  \item The final recommended hyperparameter setting is calculated by taking the best 5 percent of all 
  SMBO iterations regarding the chosen performance measure and then calculating the average of each hyperparameter of these iterations , which is rounded in case of \textit{mtry} and node size. \color{black}
  \end{itemize}
  
 \subsubsection*{Installation and execution}
  
 In the following one can see a typical example of the execution of the algorithm.  The dataset \textit{monks-problem-1} is taken from OpenML. \color{black} 
 Execution time can be estimated beforehand with the function \texttt{estimateTimeTuneRanger}  which trains a random forest with default values, multiplies the training time by the number of iterations and adds 50 for the training and prediction time of surrogate models. \color{black} The function \texttt{tuneRanger}  then \color{black} executes the tuning algorithm:

  \footnotesize
\begin{knitrout}
\definecolor{shadecolor}{rgb}{0.969, 0.969, 0.969}\color{fgcolor}\begin{kframe}
\begin{alltt}
\hlkwd{library}\hlstd{(tuneRanger)}
\hlkwd{library}\hlstd{(mlr)}
\hlkwd{library}\hlstd{(OpenML)}
\hlstd{monk_data_1} \hlkwb{=} \hlkwd{getOMLDataSet}\hlstd{(}\hlnum{333}\hlstd{)}\hlopt{$}\hlstd{data}
\hlstd{monk.task} \hlkwb{=} \hlkwd{makeClassifTask}\hlstd{(}\hlkwc{data} \hlstd{= monk_data_1,} \hlkwc{target} \hlstd{=} \hlstr{"class"}\hlstd{)}

\hlcom{# Estimate runtime}
\hlkwd{estimateTimeTuneRanger}\hlstd{(monk.task)}
\hlcom{# Approximated time for tuning: 1M 13S}
\hlkwd{set.seed}\hlstd{(}\hlnum{123}\hlstd{)}
\hlcom{# Tuning}
\hlstd{res} \hlkwb{=} \hlkwd{tuneRanger}\hlstd{(monk.task,} \hlkwc{measure} \hlstd{=} \hlkwd{list}\hlstd{(multiclass.brier),} \hlkwc{num.trees} \hlstd{=} \hlnum{1000}\hlstd{,}
  \hlkwc{num.threads} \hlstd{=} \hlnum{2}\hlstd{,} \hlkwc{iters} \hlstd{=} \hlnum{70}\hlstd{,} \hlkwc{iters.warmup} \hlstd{=} \hlnum{30}\hlstd{)}
\hlstd{res}
\hlcom{# Recommended parameter settings: }
\hlcom{#   mtry min.node.size sample.fraction}
\hlcom{# 1    6             2       0.8988154}
\hlcom{# Results: }
\hlcom{#   multiclass.brier exec.time}
\hlcom{# 1      0.006925637    0.2938}
\hlcom{#}
\hlcom{# Ranger Model with the new tuned hyperparameters}
\hlstd{res}\hlopt{$}\hlstd{model}
\hlcom{# Model for learner.id=classif.ranger; learner.class=classif.ranger}
\hlcom{# Trained on: task.id = monk_data; obs = 556; features = 6}
\hlcom{# Hyperparameters: num.threads=2,verbose=FALSE,respect.unordered.factors=order,mtry=6,min.node.size=2,}
\hlcom{# sample.fraction=0.899,num.trees=1e+03,replace=FALSE}
\end{alltt}
\end{kframe}
\end{knitrout}
\normalsize

We also performed a benchmark with 5 times repeated 5-fold cross-validation to compare it with a standard random forest with 1000 trees trained with \texttt{ranger} on this dataset. As can be seen below, we achieved an improvement of 0.014 in the error rate and 0.120 in the Brier score. 

  \footnotesize
\begin{knitrout}
\definecolor{shadecolor}{rgb}{0.969, 0.969, 0.969}\color{fgcolor}\begin{kframe}
\begin{alltt}
  \hlcom{# little benchmark}
  \hlstd{lrn} \hlkwb{=} \hlkwd{makeLearner}\hlstd{(}\hlstr{"classif.ranger"}\hlstd{,} \hlkwc{num.trees} \hlstd{=} \hlnum{1000}\hlstd{,} \hlkwc{predict.type} \hlstd{=} \hlstr{"prob"}\hlstd{)}
  \hlstd{lrn2} \hlkwb{=} \hlkwd{makeLearner}\hlstd{(}\hlstr{"classif.tuneRanger"}\hlstd{,} \hlkwc{num.threads} \hlstd{=} \hlnum{1}\hlstd{,} \hlkwc{predict.type} \hlstd{=} \hlstr{"prob"}\hlstd{)}
  \hlkwd{set.seed}\hlstd{(}\hlnum{354}\hlstd{)}
  \hlstd{rdesc} \hlkwb{=} \hlkwd{makeResampleDesc}\hlstd{(}\hlstr{"RepCV"}\hlstd{,} \hlkwc{reps} \hlstd{=} \hlnum{5}\hlstd{,} \hlkwc{folds} \hlstd{=} \hlnum{5}\hlstd{)}
  \hlstd{bmr} \hlkwb{=} \hlkwd{benchmark}\hlstd{(}\hlkwd{list}\hlstd{(lrn, lrn2), monk.task, rdesc,} \hlkwc{measures} \hlstd{=} \hlkwd{list}\hlstd{(mmce, multiclass.brier))}
  \hlstd{bmr}
  \hlcom{# Result}
  \hlcom{#    task.id          learner.id mmce.test.mean multiclass.brier.test.mean}
  \hlcom{# 1 monk_data     classif.ranger     0.01511905                  0.1347917}
  \hlcom{# 2 monk_data classif.tuneRanger     0.00144144                  0.0148708}
\end{alltt}
\end{kframe}
\end{knitrout}
  \normalsize
  
   \subsubsection*{Further parameters}
  
  \color{black}
 In the main function \texttt{tuneRanger} there are several parameters that can be changed. The first argument is the task, that has to be created 
  via the \textbf{mlr} functions \texttt{makeClassifTask} or \texttt{makeRegrTask}. The argument \texttt{measure} has to be a list of the chosen 
 measures that should be optimized, possible measures can be found with \texttt{listMeasures} or in the online tutorial of \textbf{mlr}. The argument   
 \texttt{num.trees} is the number of trees that are trained, \texttt{num.threads} is the number of
  cpu threads that should be used by \textbf{ranger}, \texttt{iters} specifies the number of iterations and \texttt{iters.warmup} 
 the number of warm-up  steps \color{black} for the initial design. 
The argument \texttt{tune.parameters} can be used to specify manually a list of the tuned parameters. \color{black}
 The final recommended hyperparameter setting (average of the best 5 percent of the iterations) 
 is used to train a RF model, which can be accessed via the list element \texttt{model} in the final outcome. 

  \subsection{Benchmark study}
  
 We now compare our new R package \textbf{tuneRanger} with different software implementations with tuning procedures for random forest 
 regarding performance and execution time in a benchmark study on 39 datasets. 
  
  \subsubsection{Compared algorithms}
  We compare different algorithms in our benchmark study: 
  \begin{itemize}
\item Our package \textbf{tuneRanger} is used with its default settings (30 warm-up  steps \color{black} for the initial design, 70 iterations, tuning of the parameters \textit{mtry}, node size and sample size, sampling without replacement)  that were set before executing the benchmark experiments. \color{black} The only parameter of the function that is varied is the performance measure that has to be optimized. We do not only consider the default performance measure Brier score (\textit{\textbf{tuneRangerBrier}}) but also
examine the versions that optimize the mean missclassification error (MMCE) (\textit{\textbf{tuneRangerMMCE}}), the AUC (\textit{\textbf{tuneRangerAUC}}) and the 
logarithmic loss (\textit{\textbf{tuneRangerLogloss}}). 
Moreover, to examine if only tuning \textit{mtry} is enough we run the same algorithms with only tuning the \textit{mtry} parameter.
 \item The three tuning implementations of the R packages \textit{\textbf{mlrHyperopt}}, \textit{\textbf{caret}} and \textit{\textbf{tuneRF}} that are described 
  in Section~\ref{sec:existtune} are executed with their default setting. We did not include \textit{\textbf{mlrHyperopt}} with other performance measures because we expect similar performance as with \textit{\textbf{tuneRanger}} but very long runtimes. \color{black}
 \item The standard RF algorithm as implemented in the package \textbf{ranger} with default settings  and without tuning \color{black} is used as reference, to see  the improvement to \color{black} the default algorithm. We use \textbf{ranger} instead of the \textbf{randomForest} package, because it is faster due to the possibility of parallelization on 
 several cores \citep{Wright2017}. 
   \end{itemize}
   
   For each method, 2000 trees are used to train the random forest. Whenever possible (for \textbf{tuneRanger}, \textbf{mlrHyperopt},  
\textbf{caret} and the default \textbf{ranger} algorithm) we use parallelization with 10 CPU cores with the
   help of the \textbf{ranger} package (trees are grown in parallel). 
  
  \subsubsection{Datasets, runtime and evaluation strategy}
  
  The benchmark study is conducted on datasets from OpenML \citep{Vanschoren2013}. We use the \textbf{OpenML100} benchmarking suite 
  \citep{Bischl20172} and download it via the \textbf{OpenML} R package \citep{Casalicchio2017}. 
For classification we only use datasets that have a binary target and no missing values, which leads to a collection of 39 datasets.  More details about these datasets such as the number of observations and variables can be found in \cite{Bischl20172}. \color{black}
  We classify the datasets into small and big by using the \texttt{estimateTimeTuneRanger} function of \textbf{tuneRanger} with 10 cores. 
  If the estimated runtime is less than 10 minutes, the dataset is classified as small, otherwise it is classified as big. This results in 26 small datasets, 13 big datasets. 
  
  For the small datasets we perform a 5-fold cross-validation and repeat it 10 times and for the  big we just perform a 5-fold cross-validation. 
  We compare the algorithms by the average of the evaluation measures mean missclassification error (MMCE), AUC, Brier score and logarithmic loss. 
  Definitions and a good comparison between measures for classification can be found in \citet{Ferri2009}. 
 In case of error messages of the tuning algorithms (on 2 datasets for \textit{\textbf{mlrHyperopt}}, on 4 datasets for \textit{\textbf{caret}} and on 3 datasets  
 for \textit{\textbf{tuneRF}}), the worst 
  result of the other algorithms are assigned to these algorithms, if it fails in more than 20 \% of the cross-validation iterations, 
   otherwise we impute missing values by the average of the rest of the results as proposed by \citet{Bischl2013}.

  \subsubsection{Results}
  \label{sec:results}
  
First, to identify possible outliers, we display the boxplots of the differences between the performances of the algorithms and the \textbf{\textit{ranger default}}. 
  Afterwards average results and ranks are calculated and analyzed. We denote the compared algorithms as 
  \textit{\textbf{tuneRangerMMCE}}, \textit{\textbf{tuneRangerAUC}}, \textit{\textbf{tuneRangerBrier}},  \textit{\textbf{tuneRangerLogloss}}, 
  \textit{\textbf{hyperopt}},  \textit{\textbf{caret}}, \textit{\textbf{tuneRF}} and \textit{\textbf{ranger default}}.
  
\subsubsection*{Outliers and boxplots of differences}
\begin{figure}[htb!]
\includegraphics[width=\maxwidth]{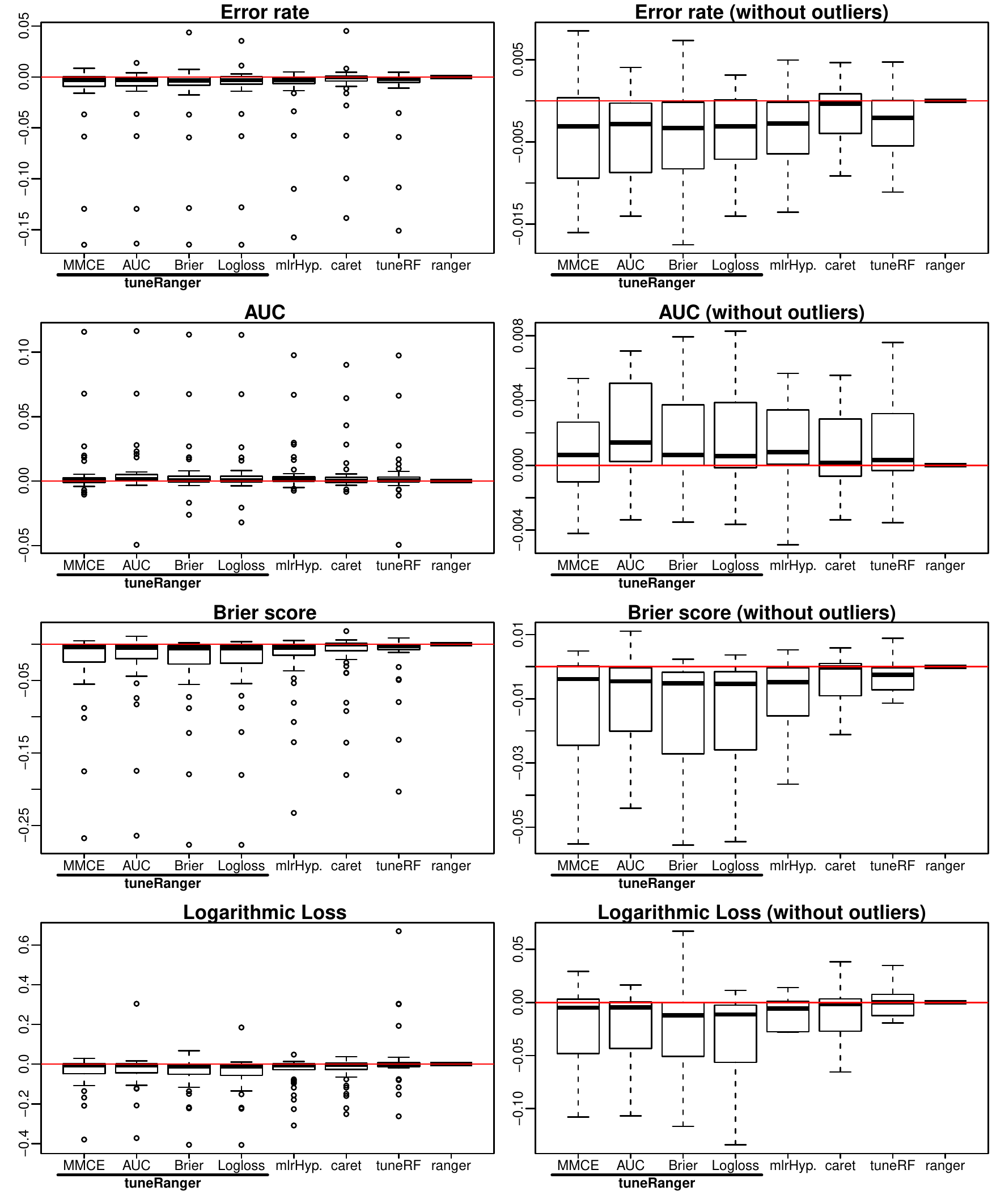} \caption[Boxplots of performance differences to \textbf{\textit{ranger default}}]{Boxplots of performance differences to \textbf{\textit{ranger default}}. On the left side the boxplots with outliers are depicted and on the right side the same plots without outliers. For the error rate, the Brier score and the logarithmic loss, low values are better, while for the AUC, high values are preferable. If the tuned measure equals the evaluation measure the boxplot is displayed in grey.}\label{fig:boxplots}
\end{figure}

In Figure \ref{fig:boxplots} on the left side, the boxplots of the differences between the performances of the algorithms and the \textbf{\textit{ranger default}} with outliers are depicted. 

We can see that there are two datasets, for 
which the performance of default random forest is very bad: the error rate is by around 0.15 higher than for all the other algorithms (other evaluation measures behave similarly). For these two datasets it is essential to tune \textit{mtry}, which is done by all tuning algorithms. The first dataset (called \textit{monks-problems-2} in OpenML) is an artificial dataset which has 6 categorical predictor variables. If two of them take the value 2 the binary outcome is 1, otherwise 0. Setting \textit{mtry} to the default 2 (or even worse to the value 1) leads to wrongly partitioned trees: the dependence structure between the categorical variables cannot be detected perfectly as sometimes the wrong variables are used for the splitting (since one of the two variables that were randomly chosen in the splitting procedure has to be used for the considered split). On the other hand, setting \textit{mtry} to 6 leads to nearly perfect predictions as the dependence structure can be identified perfectly. 
The second dataset with bad performance of the default random forest is called \textit{madelon} and has 20 informative and 480 non-informative predictor variables. The default \textit{mtry} value of 22 is much too low, because too many non-informative predictor variables are chosen in the splits. The tuning algorithms choose a higher \textit{mtry} value and get much better performances on average. 
For \textit{\textbf{tuneRF}} we have 3 outliers regarding the logarithmic loss. This tuning algorithm tends to yield clear-cut predictions with probabilities near 0 and 1 for these datasets, which lead to the bad performance regarding the logarithmic loss in these cases.
  
  \subsubsection*{Average results and ranks}
  
  The average results for the 39 datasets can be seen in Table~\ref{avg_small} and the average ranks in Table~\ref{rank_small}. The ranks are given 
   for each measure and each dataset separately from 1 (best) to 8 (worst) and then averaged over the datasets. 
  Moreover, on the right side of Figure~\ref{fig:boxplots} we can see the boxplots of the performance differences to the \textbf{\textit{ranger default}} 
  without the outliers, which give an impression about the distribution of the performance differences. 

\begin{table}[ht]
\centering
\begin{tabular}{rrrrrr}
  \hline
 & MMCE & AUC & Brier score & Logarithmic Loss & Training Runtime \\ 
  \hline
tuneRangerMMCE & 0.0923 & 0.9191 & 0.1357 & 0.2367 & 903.8218 \\ 
  tuneRangerAUC & 0.0925 & 0.9199 & 0.1371 & 0.2450 & 823.4048 \\ 
  tuneRangerBrier & 0.0932 & 0.9190 & 0.1325 & 0.2298 & 967.2051 \\ 
  tuneRangerLogloss & 0.0936 & 0.9187 & 0.1330 & 0.2314 & 887.8342 \\ 
  mlrHyperopt & 0.0934 & 0.9197 & 0.1383 & 0.2364 & 2713.2438 \\ 
  caret & 0.0972 & 0.9190 & 0.1439 & 0.2423 & 1216.2770 \\ 
  tuneRF & 0.0942 & 0.9174 & 0.1448 & 0.2929 & 862.9917 \\ 
  ranger default & 0.1054 & 0.9128 & 0.1604 & 0.2733 & 3.8607 \\ 
   \hline
\end{tabular}
\caption{Average performance results of the different algorithms.} 
\label{avg_small}
\end{table}
\begin{table}[ht]
\centering
\begin{tabular}{rrrrrr}
  \hline
 & Error rate & AUC & Brier score & Logarithmic Loss & Training Runtime \\ 
  \hline
tuneRangerMMCE & 4.19 & 4.53 & 4.41 & 4.54 & 5.23 \\ 
  tuneRangerAUC & 3.77 & 2.56 & 4.42 & 4.22 & 4.63 \\ 
  tuneRangerBrier & 3.13 & 3.91 & 1.85 & 2.69 & 5.44 \\ 
  tuneRangerLogloss & 3.97 & 4.04 & 2.64 & 2.23 & 5.00 \\ 
  mlrHyperopt & 4.37 & 4.68 & 4.74 & 4.90 & 7.59 \\ 
  caret & 5.50 & 5.24 & 6.08 & 5.51 & 4.36 \\ 
  tuneRF & 4.90 & 5.08 & 5.44 & 6.23 & 2.76 \\ 
  ranger default & 6.17 & 5.96 & 6.42 & 5.68 & 1.00 \\ 
   \hline
\end{tabular}
\caption{Average rank results of the different algorithms.} 
\label{rank_small}
\end{table}

We see that the differences are small, although on average all algorithms perform better than the \textbf{\textit{ranger default}}. 
The best algorithm is on average by around 0.013 better regarding the error rate (MMCE) and by 0.007 better in terms of the AUC. These small differences are not very surprising as the results by \citet{Probst2018} 
suggest that random forest is one of the machine learning algorithms that are less tunable.  


On average the \textit{\textbf{tuneRanger}} methods outperform \textit{\textbf{ranger}} with default settings for all measures. Also tuning the specific measure does 
on average always provide the best results among all algorithms among the \textit{\textbf{tuneRanger}} algorithms. 
It is only partly true if we look at the ranks: \textit{\textbf{tuneRangerBrier}} has the best average rank for 
the error rate, not \textit{\textbf{tuneRangerMMCE}}.
\textit{\textbf{caret}} and \textit{\textbf{tuneRF}} are on average better than \textbf{\textit{ranger default}} (with the exception of the logarithmic loss), but are clearly outperformed by most of the \textit{\textbf{tuneRanger}} methods for most of the measures.
\textit{\textbf{mlrHyperopt}} is quite competitive and achieves comparable performance to the 
\textit{\textbf{tuneRanger}} algorithms. This is not surprising as it also uses \textbf{mlrMBO} for tuning like \textbf{tuneRanger}. Its main disadvantage is the runtime. It uses 25 iterations in the SMBO procedure compared to 100 in our case, which makes it a bit faster for smaller datasets. But for bigger datasets it takes longer as, unlike \textbf{tuneRanger}, it does not use the out-of-bag method for internal evaluation but 10-fold cross-validation, which takes around 10 times longer.

\begin{figure}[htb!]
\includegraphics[width=\maxwidth]{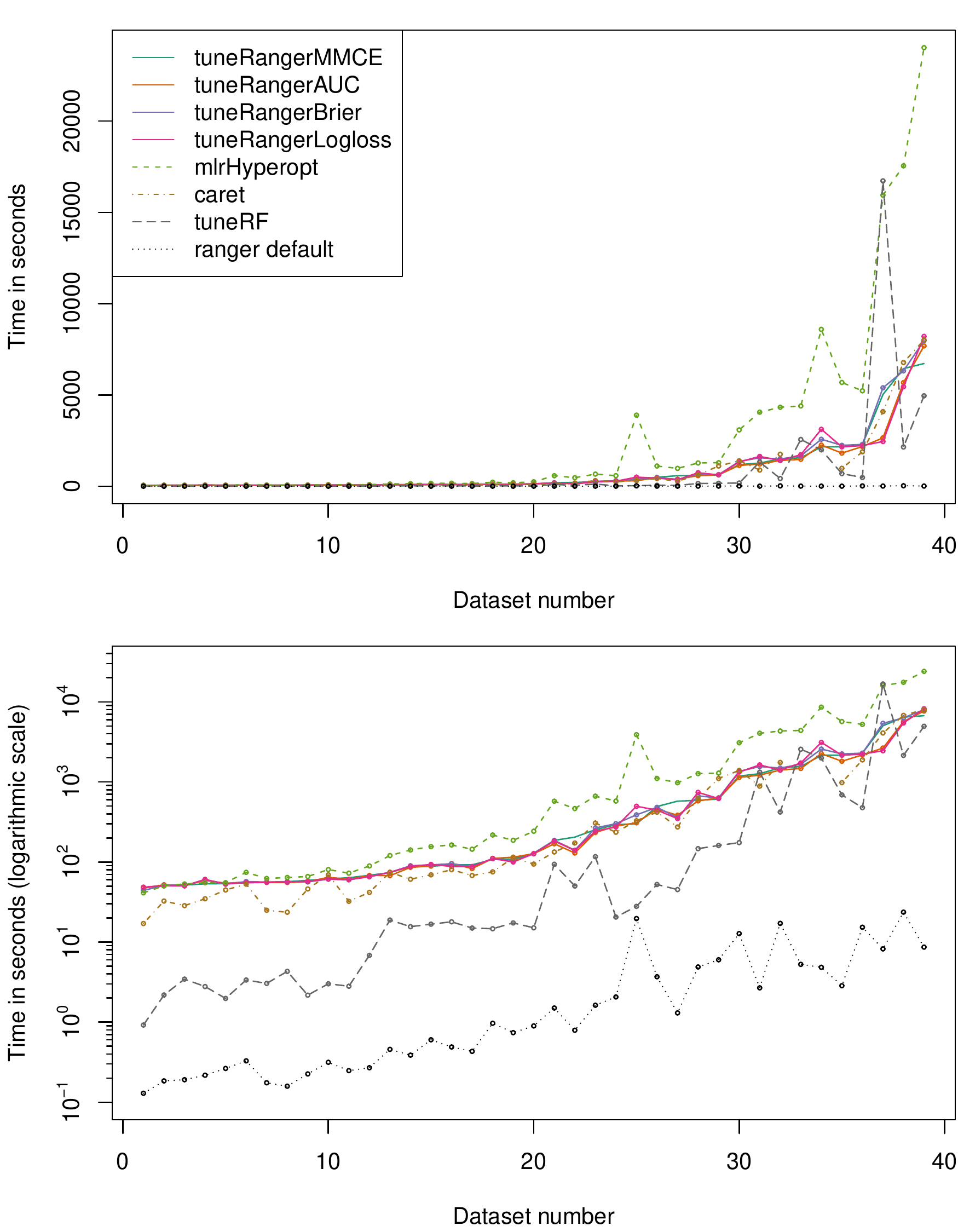} \caption[Average runtime of the different algorithms on different datasets (upper plot]{Average runtime of the different algorithms on different datasets (upper plot: unscaled, lower plot: logarithmic scale). The datasets are ordered by the average runtime of the \textit{\textbf{tuneRangerMMCE}} algorithm.}\label{fig:runtime}
\end{figure}

Figure \ref{fig:runtime} displays the average runtime  in seconds \color{black} for the different algorithms and different datasets. The datasets are ordered by the runtime of the \textit{\textbf{tuneRangerMMCE}} algorithm. For most of the datasets, \textit{\textbf{tuneRF}} is the fastest tuning algorithm, although there is one dataset for which it takes longer than all the other datasetes. The runtime of \textit{\textbf{mlrHyperopt}} is similar to the runtime of \textit{\textbf{tuneRanger}} for smaller datasets, but when
runtime increases it gets worse and worse compared to the \textit{\textbf{tuneRanger}} algorithms. For this reason we claim that \textbf{tuneRanger} is preferable especially for bigger datasets, when runtime also plays a more important role. 

To examine if tuning only \textit{mtry} could provide comparable results to tuning the parameters \textit{mtry}, node size and sample size all together we run the \textit{\textbf{tuneRanger}} algorithms with only tuning \textit{mtry}. The results show that tuning 
the node size and sample size provides on average a valuable improvement. On average the error rate (MMCE) improves by 0.004, the 
AUC by 0.002, the Brier score by 0.010 and the logarithmic loss by 0.014 when tuning all three parameters. 

\FloatBarrier

  \section{Conclusion and Discussion}
  \label{sec:conclusion}
  
The RF algorithm has several hyperparameters that may influence its performance. The number of trees should be set high: the higher the number of trees, the better the results in terms of performance and precision of variable importances. However, the improvement obtained by adding trees diminishes as more and more trees are added. The hyperparameters \textit{mtry}, sample size and node size are the parameters that control the  randomness of the RF. They should be set to achieve a reasonable strength of the single trees without too much correlation between the trees (bias-variance trade-off). 
Out of these parameters, \textit{mtry} is most influential both according to the literature and in our own experiments. The best value of \textit{mtry} depends on the number of
 variables that  are related to \color{black} the outcome. Sample size and node size have a minor influence on the performance but are worth tuning in many cases as we also showed empirically in our benchmark experiment. As far as the splitting rule is concerned, there exist several alternatives to the standard RF splitting rule, for example those used in conditional inference forests \citep{hothorn2006unbiased} or extremely randomized trees \citep{Geurts2006}. 
 
The literature on RF cruelly lacks systematic large-scale comparison studies on the different variants and values of hyperparameters. It is especially scarce as far as the impact on variable importance measures is concerned. This is all the more regrettable given that a large part of the data analysts using RF pay at least as much attention to the output variable importances as to the output prediction rule. Beyond the special case of RF, literature on computational methods tends to generally focus on the development of new methods as opposed to comparison studies investigating existing methods. As discussed in \cite{boulesteix2018necessity}, computational journals often require the development of novel methods as a prequisit for publication. Comparison studies presented in papers introducing new methods are often biased in favor of these new methods---as a result of the publication bias and publication pressure. As a result of this situation, {\it neutral} comparison studies as defined by \cite{boulesteix2017towards} (i.e., focusing on the comparison of existing methods rather than aiming at demonstrating the superiority of a new one, and conducted by authors who are as a group approximately equally competent on all considered methods) are important but rare.

The literature review presented in this paper, which is to some extent disappointing in the sense that clear guidance is missing, leads us to make a plea for more studies investigating and comparing the behaviors and performances of RF variants and hyperparameter choices. Such studies are, in our opinion, at least as important as the development of further variants that would even increase the need for comparisons.
 
In the second part of the paper, different tuning methods for random forest are compared in a benchmark study. The results and previous studies show that tuning random forest can 
improve the performance although the effect of tuning is much smaller than for other machine learning algorithms such as support vector machines \citep{Mantovani2015}. Out of 
existing tuning algorithms, we suggest to use sequential model-based optimization (SMBO) to tune the parameters \textit{mtry}, sample size and node size simultanously. Moreover, 
the out-of-bag predictions can be used for tuning. This approach is faster than, for example, cross-validation. The whole procedure is implemented in the 
package \textbf{tuneRanger}. This package allows users to choose the specific measure that should be minimized (e.g., the AUC in case of classification). 
In our benchmark study it achieved on average better performances than the standard random forest and other software that implement tuning 
for random forest, while the fast {\bf tuneRF} function from the package {\bf randomForest} can be recommended if computational speed is an issue.

\FloatBarrier

\end{document}